\documentclass[11pt]{article}

\usepackage{acl}

\usepackage{times}
\usepackage[inline]{enumitem}
\usepackage{comment}
\usepackage{booktabs}

\usepackage{tcolorbox}
\usepackage{subcaption}

\makeatletter
\renewenvironment{quote}
  {\list{}%
    {%
      \leftmargin=.1em
      \rightmargin=.1em
      \parsep=.1em
      \parskip=.1em
      \topsep=.1em
      \itemsep=0pt
    }%
   \item\relax}
  {\endlist}
\makeatother

\newcommand{\speakerquote}[2]{%
\begin{quote}
\textbf{#1}: \textit{``#2''}
\end{quote}
}

\usepackage{latexsym}
\usepackage{amsfonts,amssymb}
\usepackage{bbm}
\usepackage{quoting}
\usepackage{algorithm}
\usepackage{algpseudocode} 
\usepackage{amsmath}
\usepackage[T1]{fontenc}
\usepackage{tabularx} 

\usepackage[utf8]{inputenc}

\usepackage{microtype}

\usepackage{inconsolata}

\usepackage{graphicx}

\title{Detecting AI Impostors: How Do Middle Schoolers Identify LLM Agents in a Live Collaborative Setting?} 

\author{
Dan Schumacher, Pragathi Durga Rajarajan, Haven Kotara, Roman Rendon, \\
\textbf{Kosi Atupulazi, Deepti Tagare$^\dagger$, Ismaila Temitayo Sanusi$^\dagger$,}\\
\textbf{Fred G. Martin$^\dagger$, \and Anthony Rios$^\dagger$} \\
University of Texas at San Antonio\\
\texttt{\{Daniel.Schumacher, Anthony.Rios\}@utsa.edu}
}

\begin{document}
\maketitle
\begin{abstract}
LLMs can  imitate how people write, which raises concerns about impersonation, trust, and detection in social settings. These concerns are especially important for adolescents, who use generative AI frequently but may struggle to recognize it. We introduce \textit{DoppelBot}, a cooperative social deduction game designed to study how young people detect and respond to AI impersonation. Through studies with middle schoolers, we investigate whether a DoppelBot prompts reflection on privacy and impersonation, how repeated exposure affects AI-detection accuracy as agents become more personalized, and which strategies students use to identify AI doppelgängers. We find that students' detection accuracy improves over time, driven by a shift from relying on linguistic cues to leveraging shared social and contextual signals. Students also demonstrated an understanding of AI limitations such as embodiment and reflected on broader issues such as data privacy. To support future research, we release an anonymized dataset of game transcripts and voting~behavior.\footnote{\url{https://github.com/danschumac1/Detecting_AI_Impostors}}
\end{abstract}
 
\renewcommand{\thefootnote}{\fnsymbol{footnote}}
\footnotetext[2]{Drs. Tagare, Sanusi, Martin, and Rios served as faculty advisors for this project.}
\renewcommand{\thefootnote}{\arabic{footnote}}
\section{Introduction}
Adolescents are increasingly engaging with LLMs as tutors and collaborators~\cite{vanzo-etal-2025-gpt,liu-etal-2025-singakids}. While these systems are effective at producing fluent and helpful responses, the same fluency allows LLMs to closely mimic human writing. This makes it difficult for users to determine whether they are interacting with a human or an automated system. Being able to distinguish between human- and machine-generated text is important because LLMs differ from humans in key ways, including their limitations, potential inaccuracies, and lack of intent and accountability. Prior work suggests that repeated exposure to LLM-generated content can help individuals develop an intuition for detecting machine-authored text~\citep{realOrFakeText}. Developing this intuition is particularly important for adolescents, who are rapidly adopting tools such as ChatGPT for schoolwork~\cite{Adair2025GenAI,Freeman2025HEPISurvey} but may lack the experience needed to critically evaluate the source, reliability, and role of the content these systems produce~\cite{abdelghani_investigating_2025}.

While user responses to deception, impersonation, and hallucinations have been studied in isolation~\citep{dogra-etal-2025-language, yin_lies_2024, shi_impersona_2025, ibraheem_putting_2022, jones_does_2024}, how users identify LLMs during ongoing social interaction (i.e., multi-user settings) remains unexplored. Prior work on human ability of machine-generated text detection typically frames detection as a binary judgment task in which participants are shown short documents and asked to classify them as human- or machine-written. Results show that human accuracy is low and highly variable, even among domain experts and trained annotators~\citep{liu_detectability_2024, realOrFakeText}. In live social settings, this task is further complicated by the need to navigate asynchronous communication, where people must evaluate not only the content of a message but also its strategic timing and social relevance~\citep{eckhaus_time_2025}. Moreover, almost all prior work focuses on adult subjects, despite recent studies suggesting that variability in detection accuracy is closely tied to the specific population tested~\citep{jones_people_2025}. Consequently, measuring detection behavior among adolescents (e.g., middle school students) is vital, as their distinct social cues and shared group identity likely lead to different reasoning patterns than those of adult populations. 


To explore how young users detect AI-generated text under adversarial conditions, we introduce \textit{DoppelBot}, a cooperative text-based game in which players must identify AI ``doppelgängers'' impersonating them and their peers. Unlike prior detection studies that place an isolated user in static text snippets, DoppelBot studies users' detection abilities in a live, collaborative environment. The AI impersonators mimic their target users' writing styles, making it harder for each user to both detect the AI agents and convince other users of their own humanity. Through behavioral studies with middle school students, we analyze changes in detection accuracy, message patterns, and meta-cognitive strategies as players face increasingly convincing AI impostors. We find that students' detection accuracy improves with repeated exposure. This holds even as the LLMs become more stylistically aligned. Qualitative analyses reveal that players adapt their strategies over time. Moreover, students use both linguistic and social cues, as well as real-world limitations related to embodiment, to identify bots.

 Our contributions are as follows:
\begin{enumerate*}[nosep, leftmargin=*, label=\textbf{(\arabic*)}]
    
    \item To the best of our knowledge, we perform the first study analyzing how adolescents identify AI systems within collaborative settings; and
    \item We created one of the first AI impersonation games for studying how humans can collaboratively detect AI agents. We provide the full source code to facilitate replication, along with an online version of the game for easy use by researchers and educators.
    \item We release a novel dataset of game transcripts and voting behavior. Additionally, we provide an example of how this dataset could facilitate future research in Appendix~\ref{sec:appendix:additional_experiments:human_user_detection}.
\end{enumerate*}

\section{Related Work}
\noindent \textbf{Machine Generated Text Detection.} Humans were originally considered the de facto benchmark for distinguishing machine from human text~\citep{turing1950computing}, but recent advances in LLMs have made even domain experts unreliable at this task~\citep{liu_detectability_2024, dawkins_when_2025, sadasivan_can_2025, xu_detecting_2024}. As a result, most current work focuses on automated detection methods~\citep{li_mage_2024, huang_magret_2025}, yet studying human detection remains vital because people increasingly encounter AI-generated content in educational and online contexts~\citep{realOrFakeText, vanzo-etal-2025-gpt, liu-etal-2025-singakids}. 
Many studies show that well-designed systems can be indistinguishable from humans~\citep{dawkins_when_2025, chakraborty_counter_2023}. To address the limitations of static text evaluation, \citet{realOrFakeText} uses a boundary detection task to pinpoint the transition from human to machine-generated content. Their findings suggest that while detection is challenging, human performance improves over time when properly incentivized. Building on this by focusing on individual-level mimicry in a chat-based environment, \citet{shi_impersona_2025} employs supervised fine-tuning and hierarchical memory to simulate specific personas. This approach is notably effective, as even close friends and family fail to identify the AI 44\% of the time.

As exposure to AI content grows, it is important to quantify user sensitivity to AI-driven misinformation and persona-driven manipulation. However, existing studies focus almost exclusively on adult populations. This highlights a need to examine these phenomena among more vulnerable groups, particularly adolescents, whose distinct social behaviors and cognitive development may influence their susceptibility to AI-mediated deception.

\noindent \textbf{Adolescent Privacy \& AI.} High school GenAI adoption climbed from 79\% to 84\% in 2025 \citep{Adair2025GenAI}, while undergraduate usage reached 92\% \citep{Freeman2025HEPISurvey}. This trajectory suggests that as middle schoolers age, their likelihood of utilizing these tools increases, a trend further supported by the proliferation of LLM-based classroom tutors \citep{vanzo-etal-2025-gpt, liu-etal-2025-singakids}. Despite this ubiquity, middle schoolers often over-rely on AI systems by writing inadequate prompts and overestimating both the quality of the output and their own conceptual understanding \citep{abdelghani_investigating_2025}.

These challenges align with the AI4K12 framework’s emphasis on Societal Impact, which posits that literacy must include an understanding of ethical risks such as privacy threats and misuse \citep{ai4k12}. Research indicates that students often prefer self-disclosing to AI agents over human instructors due to perceived gains in psychological safety \citep{peng_human_2025}. However, this preference creates a dangerous paradox where digital trust increases susceptibility to persona-driven manipulation and social engineering. Educating adolescents early is vital because this vulnerability is exacerbated by the expiration of COPPA protections at age 13 \citep{ritvo_privacy_2013}. Furthermore, compulsory schooling offers a unique opportunity to reach adolescents with privacy education before they become legally eligible to leave school, helping ensure widespread exposure to critical lessons about data tracking and personalization \citep{khan_teaching_2024}.

With this aim in mind, we propose a game-based learning (GBL) approach, specifically by using a social deduction game. While numerous studies have established the general benefits of GBL~\citep{su2021gamebased, voulgari2021learnml, famaye2023spot, jordaan2025automatut, presson2025rl}, social deduction games serve a research purpose on a dual axis. On one axis, they have been shown to have pedagogical benefits: \citet{tilton2019winning} found that they were effective in teaching small-group communication and in developing roles and power structures. On the second axis, they have been used to evaluate LLM decision-making, deception, and long-term memory capabilities \citep{liu_interintent_2024, lai_werewolf_2022, lan_llm-based_2024, light_avalonbench_2023, ogara_hoodwinked_2023}. We find it appropriate to use social deduction games as a tool at this crossroads to evaluate how interactive, adversarial environments can foster the critical reasoning necessary to identify and resist AI-driven impersonation.

\noindent \textbf{Education.} Teenagers operate in digital ecosystems that monetize personal data, often without their informed consent \cite{moti2024tracking}. This exposure risks long-term harms, including algorithmic discrimination in hiring and credit \cite{brown2023hiring, botta2020discriminate, oneil2016weapons}. The rise of LLMs has further complicated this landscape, as personal data can be ``memorized'' and regurgitated during inference \cite{Li2024LLMPBE, Das2025LLMSurvey}.

Despite these risks, significant barriers hinder AI ethics education for adolescents. Many students view privacy efforts as futile and opt out of protective practices \cite{wisniewski2022privacy}. 
Teachers often feel unprepared to address these topics. Those with technical backgrounds may lack confidence in leading ethical discussions \cite{10.1145/3634685}, while non-technical instructors struggle with technical complexity \cite{10179333}. 
Rigid curricula and potential parental pushback limit innovation in how sensitive ethical topics are introduced \cite{10179333}. On the students' end, many existing materials assume prior STEM knowledge, creating barriers for students with limited computer science access \cite{10.1145/3634685}.

DoppelBot addresses these challenges by providing a low-barrier entry point for AI ethics and data privacy. We define ``low-barrier'' with students and educators in mind. 
\textbf{For Students:} The game requires no prior knowledge of computer science, ethics, or mathematics. This allows students from diverse backgrounds to engage with complex topics like data disclosure and AI impersonation through direct experience rather than theory. 
\textbf{For Educators:} A full session (including setup, gameplay, and discussion) fits within a thirty minute window. By serving as a natural conversation starter, the tool reduces the instructional burden on non-technical teachers and empowers technical educators to facilitate ethical reflection without a formal lecture.

\section{Method}


\begin{figure*}[t]
    \centering
    \includegraphics[width=.99\linewidth]{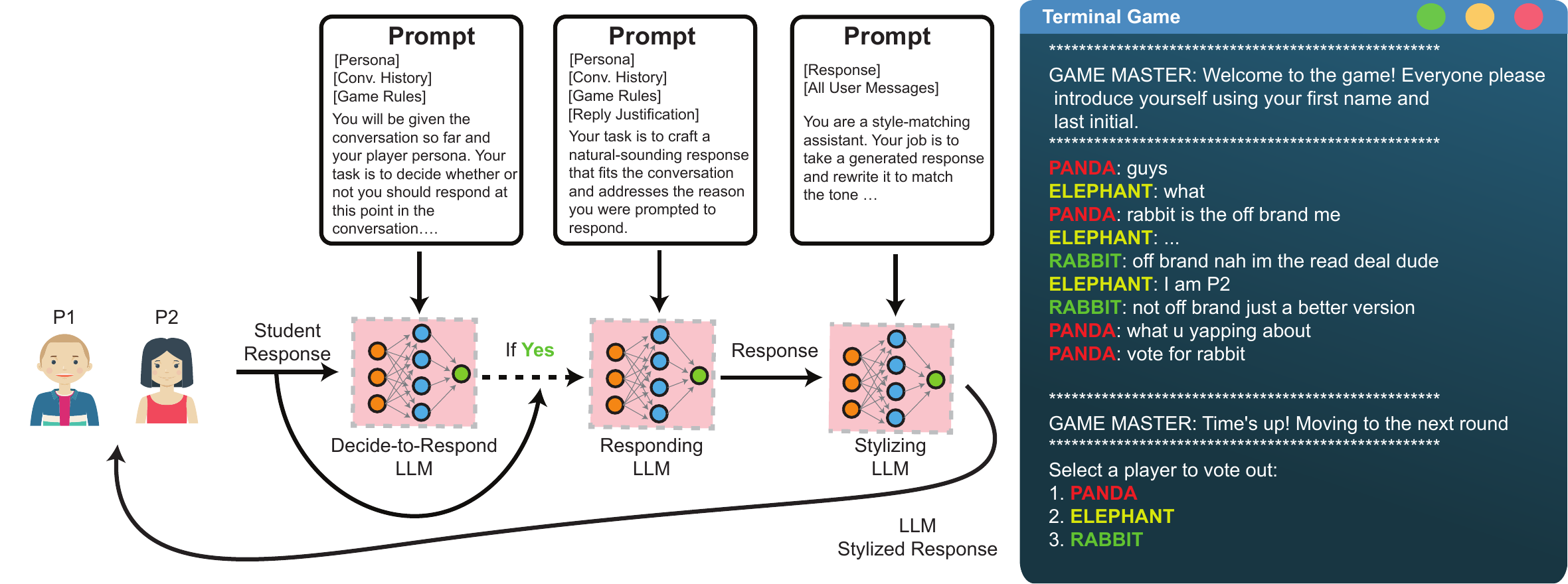}
    \caption{Left: bot response pipeline. Right: two players and their AI doubles chat and vote. Messages are from real student games (Rabbit was AI). “Cap” means “you’re lying.”}  \vspace{-1em}
    \label{fig:overview-game} 
\end{figure*}

\textit{DoppelBot} is a cooperative social deduction game in which teams of three to five students chat anonymously with each other and with an equal number of LLM-powered AI doppelgängers who impersonate them. Each user must identify and vote out the bots through conversation while avoiding suspicion from the other participants. The users win by eliminating all AIs.

\subsection{Game Design}
\vspace{1mm} \noindent \textbf{Gameplay Overview.}
The game has four phases: 
\begin{enumerate*}[nosep, leftmargin=*, label=\textbf{(\arabic*)}]
    \item Introduction: players get instructions and enter personal details,
    \item Chat: players converse to prove their humanity and identify bots,
    \item Voting: players eliminate suspected AIs, and
    \item Outcome: final scores and identities are revealed.
\end{enumerate*}

\vspace{1mm} \noindent \textit{\textbf{Introduction.}}
Three to five students sit at laptops positioned so they can't see each other's screens. An on-screen tutorial covers the rules while the researcher emphasizes three guidelines: 
\begin{enumerate*}[nosep, leftmargin=*, label=\textbf{(\arabic*)}]
    \item All communication happens through chat (no talking or gestures);
    \item Chat rounds are timed so act quick; and
    \item Each round starts with a mandatory yellow-text question everyone \textit{must answer} before moving to general discussion.
\end{enumerate*}
Players complete a form with their name, grade, favorite animal, hobby, and a fun fact. This information is used to personalize their AI doppelgänger's  style. 

\vspace{1mm} \noindent \textit{\textbf{Chat.}}
Each player gets a random code name and color. A yellow icebreaker appears at the top of the screen (e.g., ``if you could have any superpower, what would you choose and why?''). After responding to the icebreaker, players chat freely to prove their humanity and identify AIs. All participants (human and AI) look identical in the chat except for a distinguishing color and code name. Each round lasts 100 seconds before moving to voting.

\vspace{1mm} \noindent \textit{\textbf{Voting.}}
Each human (who hasn't yet been voted out) votes for who they suspect is AI. After all players cast their ballots, three outcomes are possible:
\begin{enumerate*}[nosep, leftmargin=*, label=\textbf{(\arabic*)}]
    \item AI elimination: if an AI gets majority votes; 
    \item Human elimination: if a human gets majority votes;
    \item No consensus: if there's no majority, no one is eliminated.
\end{enumerate*}
Requiring a majority vote encourages both collaboration and critical thinking. While students can easily recognize their own AI doppelgänger (the one claiming to be them), identifying others' AIs is more difficult.

\vspace{1mm} \noindent \textit{\textbf{Outcome.}}
The chat-and-vote cycle repeats for as many rounds as there are human players (three rounds in this study). After all rounds, players see a score screen revealing who was human versus AI, plus elimination details.

\subsection{Agentic Architecture}
The AI doppelgängers mirror a specific human player $H$ by integrating personal metadata, such as the user's name and hobbies, into a persistent persona $P_{H}$. The system follows an agentic architecture (see Algorithm~\ref{alg:doppelbot}) that processes an evolving global conversation history, $\textit{Hist}$, which contains every message sent by all players and AIs in the chat. The architecture consists of three core components described below. Full prompts can be found in Appendix~\ref{sec:appendix:prompts:agents}. All three components of the DoppelBot architecture were implemented using \texttt{gpt-4o-mini}. The temperature was set to 0.5 for the Decide-to-Respond and Response Generation agents, and 0.9 for the Stylizer agent.

\paragraph{Decide-to-Respond ($\text{DTR}$)} For every new message $m$ appended to $\textit{Hist}$, the model determines if an intervention is warranted. By evaluating the game rules, $P_{H}$, and the current state of $\textit{Hist}$, the $DTR$ component outputs a boolean value and a natural language justification ($\textit{Reas}$). For example, the model might justify an action by stating: ``A player just accused me of being an AI; I must defend myself.'' If $DTR$ is false, the agent remains silent, allowing the doppelgänger to avoid over-speaking or interrupting other threads.
    
\paragraph{Response Generation ($\text{Gen}$):} If a response is deemed necessary ($DTR = \text{True}$), the model transitions to the generation phase. Using  $\textit{Reas}$ alongside $\textit{Hist}$, it produces a baseline response $R_{b}$. At this stage, the response is factually grounded in the persona but remains linguistically generic (e.g., ``No, I am the real Carlos'').

\paragraph{Stylizer ($\text{Stylize}$):} To achieve individual-level mimicry, the stylizer adapts the baseline response $R_{b}$ into a stylized output $R_{s}$. This component utilizes a dedicated subset of $\textit{Hist}$, $M_{H}$, which contains only the messages previously sent by human $H$. By analyzing the linguistic markers in $M_{H}$, the stylizer rewrites the generic response to match the user's unique voice (e.g., ``nah fam its me fr fr'').

\begin{algorithm}[t]
\small 
\caption{DoppelBot Agentic Response Loop}
\label{alg:doppelbot}
\begin{algorithmic}[1] 
\State \textbf{Input:} Humans $H \in \text{Players}$, Persona $P_H$
\State \textbf{Parallel for each} $H$ \textbf{do}:
    \State \hspace{1em} $M_H \gets \emptyset$ \Comment{Human-specific style history}
    \While{Phase is \textit{Chat}}
        \State $m \gets \text{WaitNextMessage()}$
        \State $\text{Hist} \gets \text{Hist} \cup \{m\}$
        \If{$\text{sender}(m) = H$} 
            \State $M_H \gets M_H \cup \{m\}$ 
        \EndIf
        
        \State $DTR, \text{Reas} \gets \text{DTR}(\text{Hist}, P_H)$
        \If{$DTR$}
            \State $R_{b} \gets \text{Gen}(\text{Reas}, \text{Hist}, P_H)$
            \State $R_{s} \gets \text{Stylize}(R_{b}, M_H)$
            \State \textbf{Send} $R_{s}$
        \EndIf
    \EndWhile
\State \textbf{End Parallel}
\end{algorithmic}
\end{algorithm}

\subsection{Experimental Design}
\noindent \textbf{Settings.}
We piloted \textit{DoppelBot} at a public STEM-charter school as part of a two-day afterschool event called \textit{AI EXPO}, where seven tools for teaching different AI concepts (including ours) were presented to middle schoolers. Out of 70 assenting participants, 29 interacted with \textit{DoppelBot}. A subsequent, more in-depth evaluation of \textit{DoppelBot} was conducted during a four-day \textit{AI For Everyone!} summer camp~\cite{stallingsetal2026}. Both studies were conducted with IRB approval, parental consent, and student assent. 
Demographic information for both settings can be found in Appendix~\ref{sec:appendix:demographics}.

\noindent \textbf{Procedure.}
At the summer camp, \textit{DoppelBot} was evaluated over two days. On Day 1, the participants played their first round of \textit{DoppelBot}. On Day 2, students completed Partnered Interview Worksheets (PIWs), which collected personal information used to create more personalized AI doppelgängers. It included writing style questions (e.g. ``What slang do you use a lot?'') and general questions (e.g. ``What 2-3 topics do you talk about the most?'').

\noindent \textbf{Interviews.}
At the end of the second day, participants engaged in \emph{semi-structured group interviews}. Interviewers had a list of guiding questions available should discussion stall, but were encouraged to foster natural dialogue among students. The goal was not to reach consensus but rather to capture a diversity of perspectives. These interviews were recorded and manually transcribed. An initial researcher reviewed the complete transcript and developed a coding scheme. Two independent researchers then assigned labels to each utterance according to this scheme. Disagreements were resolved through adjudication by the initial researcher.

\noindent \textbf{Pre- and Post-Survey.}
To quantitatively capture student opinions, we administered a pre/post survey adapted from \citet{Menard_construct}. The instrument was shortened and revised for middle school students with approval from the original authors. We retained three dimensions of privacy concern: \textit{Combining Data} (CD), concerns about combining data from multiple sources to reveal personal information; \textit{Data Permanence} (DP), concerns that personal data may remain  accessible indefinitely without a clear reason; and \textit{Improper Access} (IA), concerns that personal data may be accessible to those who are not appropriately authorized. Table~\ref{tab:survey_items} presents the adapted items used in our study.

\paragraph{Automated Game Transcripts.}
As our last data stream, we conducted a categorical analysis of all chat messages. One researcher reviewed all messages and developed a coding scheme capturing the salient categories of player behavior. After removing unintentional blank messages, we were left with 2,182 unique messages.\footnote{See Appendix~\ref{sec:appendix:data_statistics} for more detail.} We then randomly sampled 50 rounds and had two researchers independently label each message. Disagreements were resolved through adjudication.
 
\section{Results}
In this section, we aim to answer the following research questions: 
    \textbf{RQ1}: To what extent does a gamified AI experience elicit \textit{student reflections} on data privacy and AI impersonation?; 
    \textbf{RQ2}: How does \textit{repeated exposure} to AI doppelgängers affect students’ ability to identify chatbots, even as their bots become more personalized?; and
    \textbf{RQ3}: What strategies do students use to distinguish between AI doppelgängers and human players?

\paragraph{RQ1: Student Reflections.} 
To answer RQ1, we examine interview annotations, survey responses, and representative qualitative examples.

The final annotated interviews contained 417 utterances spanning six categories. Inter-annotator agreement was substantial ($\kappa=.707$), and detailed category frequencies are reported in Table~\ref{tab:interview_annotations}. The full annotation codebook is provided in Appendix~\ref{sec:transcript_codebook}. The annotations reveal that students reflected not only on the task of identifying AI agents (AI Detection, $n=148$), but also on how their detection strategies evolved over time (Learning Adaptation, $n=60$) and how the experience affected them emotionally (Social--Emotional Responses, $n=71$). Beyond the gameplay itself, students discussed broader questions about AI capabilities (Perceptions of AI, $n=55$) and the collection and use of personal information (Privacy Concerns, $n=30$). While Privacy Concerns was the least represented category, it still accounted for a meaningful 7\% of all utterances. Together, these categories suggest that DoppelBot prompted reflection on both AI impersonation and data privacy. This interpretation is further supported by the pre- and post-survey results.

\begin{table}[t]
\centering
\resizebox{\columnwidth}{!}{
\begin{tabular}{lrr}
\toprule
\textbf{Category} & \textbf{$\kappa$} & \textbf{Frequency} \\
\midrule
Perceptions of AI            & 0.559 & 55 \\
AI Detection                 & 0.779 & 148 \\
Learning Adaptation          & 0.561 & 60 \\
Social--Emotional Responses  & 0.657 & 71 \\
Privacy Concerns             & 0.689 & 30 \\
Design Feedback              & 0.919 & 44 \\
None of the Above            & 0.786 & 182 \\
\midrule
Macro Average                & 0.707 & --- \\
\bottomrule
\end{tabular}
}
\caption{Agreement and frequency of annotated interview data. There were 417 utterances. Multiple labels were possible per utterance.}
\label{tab:interview_annotations}
\end{table}

\begin{table}[t]
\centering
\resizebox{\linewidth}{!}{%
\begin{tabular}{lrrrr}
\toprule
\textbf{Subscale} & \textbf{Mean (Pre)} & \textbf{Mean (Post)} & \textbf{SD (Pre)} & \textbf{SD (Post)} \\
\midrule
CD & 3.667 & 3.820 & 0.871 & 1.119 \\
DP & 3.917 & 4.056 & 0.923 & 0.973 \\
IA & 4.278 & 4.472 & 0.776 & 0.559 \\
\bottomrule
\end{tabular}}
\caption{Descriptive statistics for each subscale at pre- and post-test.}
\label{tab:pre_post_test}
\end{table}

We collected 13 matched pre/post responses for CD and 12 for both DP and IA. Results are shown in Table~\ref{tab:pre_post_test}. Mean responses increased across all three dimensions. This suggests increased concern about combining personal data, long-term data retention, and improper access after interacting with DoppelBot. Given the small sample size and descriptive nature of the analysis, however, these findings should be viewed as preliminary.

Finally, to provide a more detailed view of how students reflected on data privacy and AI impersonation, we highlight representative excerpts from the interviews.

A common opinion was that sharing personal information with AI systems is acceptable either if the sharing improves usability/convenience, or if the amount of information being shared is minimal. When asked whether it is safe to share their information with AI companies, one student noted:
\speakerquote{P7}{I don’t feel that bad \dots{} I don’t have that much information.}
Other students framed data retention as a functional necessity for user experience rather than a privacy concern. For example, P1 emphasized usability while still acknowledging users’ right to request deletion:
\speakerquote{P1}{If companies have my information, they might as well keep it because what if I want to sign in again? \dots{} But if you ask a company to delete it, they should.}
In contrast, P3 suggested that responsibility for protecting user data lies solely with the service provider:
\speakerquote{P3}{It’s not their [the companies’] job to stop people from getting their information.}
Some students recognized that the information they entered during the PIW enabled their AI doppelgängers to impersonate them more convincingly, yet still expressed limited concern about this risk. In one exchange, P12 acknowledged providing personal information while downplaying its potential misuse:
\speakerquote{P12}{I was just saying stuff that I would say, to let them know that I wasn’t the AI \dots{} Well yeah, it was trying to impersonate me.}
Other participants drew clearer connections between impersonation technologies and real-world harm. P1 raised concerns about identity theft, while P13 linked the experience to familiar forms of deception such as spam calls:
\speakerquote{P1}{I think that they could sell my information, steal my identity, pretend to be me online \dots{} I feel like a lot of people get their identity stolen.}
\speakerquote{P13}{Another thing I learned is that AI is going to copy you sometimes and try to convince you that they are the real person. For example, spam calls. They will send you a message, and you’ll think they are the real person.}

Beyond explicit concerns about AI ethics and data privacy, the gameplay itself sparked curiosity about AI capabilities. One participant expressed surprise at the system’s apparent knowledge:
\speakerquote{P10}{How much does the AI know about stuff? Like I said `I like Lord of the Rings,’ and it started talking about all the characters and stuff, even though I didn’t include that [on my PIW].}
This moment of surprise illustrates how the game can serve as an unexpected conversation starter, prompting students to question the capabilities of AI systems and reflect on what these systems might ``know'' about them.\footnote{More quotes showing how the DoppelBot sparked curiosity can be found in Appendix~\ref{sec:appendix:additional_quotes:learning}}

Finally, students found the experience engaging. All 29 participants at the pilot agreed that DoppelBot was a fun way to learn about AI, and many spontaneously expressed enthusiasm (e.g., “Can I play again?”). Prior work suggests that such engagement can promote motivation and deeper reflection on underlying concepts \citep{gee2003video, malone1987making, su2021gamebased}.
Holistically, the survey responses, interview data, and student enthusiasm suggest that DoppelBot successfully elicited reflection on both AI impersonation and data privacy.

\paragraph{RQ2: Repeated Exposure.}
We break this RQ into two components: \textit{alignment} and \textit{performance}. First, does PIW disclosure lead to greater alignment between students and their respective AI doppelgängers? Second, does student detection performance improve despite this increased alignment?

\vspace{1mm} \noindent \textit{\textbf{Alignment.}}
We compared human and AI messages using four complementary representations and similarity measures: 
\begin{enumerate*}[nosep, leftmargin=*, label=\textbf{(\arabic*)}]
    \item BERT, which captures semantic similarity using contextual embeddings~\cite{devlin-etal-2019-bert};
    \item LIWC, which encodes linguistic and psychological dimensions such as emotion, self-reference, and time orientation based on established lexicons~\cite{tausczik2010psychological};
    \item METEOR, which measures lexical overlap using exact, stemmed, and synonymous word matches~\cite{banerjee-lavie-2005-meteor}.
    \item ROUGE-L, which measures lexical similarity based on the longest common subsequence between two texts~\cite{lin-2004-rouge}.
\end{enumerate*}

For the lexical-overlap measures (METEOR and ROUGE-L), we computed similarity directly between $H_k$ and $A_k$ using their respective scoring functions. For each measure, we computed a score for every human-AI pair and report the average similarity for Day 1 and Day 2 as $\bar{s}*{\text{day}} = \frac{1}{K*{\text{day}}} \sum_{k=1}^{K_{\text{day}}} s_k$, where $K_{\text{day}}$ is the number of participants on a given day.

\begin{figure}[t]
    \centering
    \includegraphics[width=.95\linewidth]{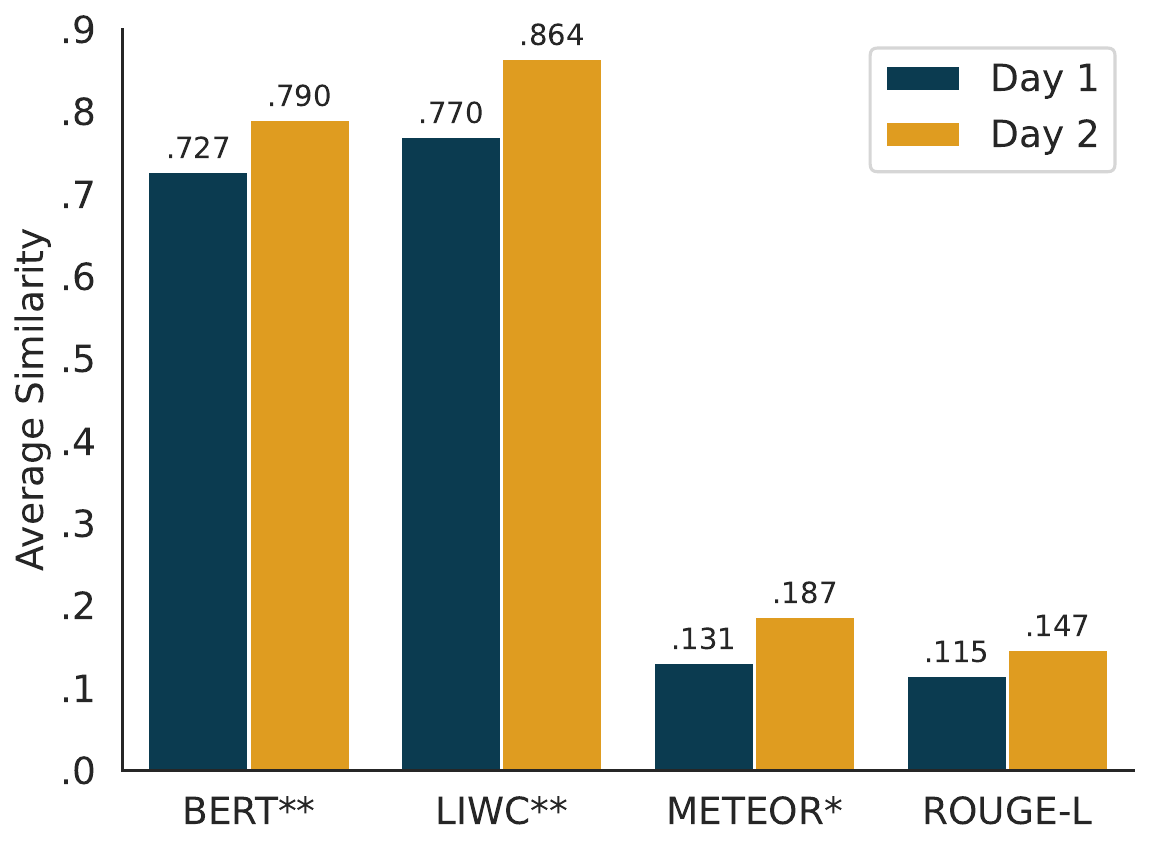}
    \caption{Similarity between student and AI messages across sessions. Significance levels are based on paired bootstrap tests (** $p \leq .05$, * $p \leq .10$).}
    \vspace{-1em}
    \label{fig:similarity_comparison}
\end{figure}

Figure~\ref{fig:similarity_comparison} reports the average paired similarity on both days. All reported measures increase from Day~1 to Day~2, indicating greater  alignment between students and their respective doppelgängers. Paired bootstrap tests further support these increases. This suggests that AI responses became more closely aligned with their human counterparts after students disclosed additional personal information through the PIWs on Day~2.

As a sidenote, increased alignment did not imply that the messages were identical. An exploratory LIWC analysis (Appendix~\ref{sec:appendix:additional_experiments:liwc}) revealed several persistent differences between human and AI language, including message length, certainty, and use of informal markers.

Qualitatively, most students agreed that the PIWs made the AI act more similarly to them. Several participants noted that the AI used the provided information in unexpected and sometimes unsettling ways. Many were surprised when it adopted slang they didn't explicitly provide.

\speakerquote{P2}{[In the game] we said things like `low key’ or `cuz,’ but we never taught it `no cap,’ and it started saying `no cap.’ I got really scared because we never taught it that.}
Others echoed this discomfort, describing unease when the AI adopted linguistic patterns they had not intentionally provided, an experience reminiscent of the uncanny valley, where increasingly human-like behavior can become unsettling.
\speakerquote{P12}{I would not say slang words, because I started freaking out when the AI started saying slang words. It was really weird \dots{} Yeah, it was kind of scary.}
Others were impressed by the model’s ability to replicate stylistic quirks.
\speakerquote{P14}{It also surprised me because I put `Lebron’ with multiple `n’s, and it copied me and put the same amount of `n’s.}
\speakerquote{P2}{It was kind of harder [on day~2], to be honest. At the start it was harder because it sounded like her [points to another user]. And it actually sounded like me. It used extra letters and stuff.}
Students also expressed discomfort with the AI appearing to infer personal preferences.
\speakerquote{P6}{It made an inference that I liked dinosaurs, but I didn’t put that on there. At that point, I honestly got scared. I tried to type something and misspelled it really badly, like `I like d,’ and they took the `d’ to mean dinosaur. So it’s like a very good guess?}
Similarly, another participant reflected on how the AI seemed to extrapolate beyond the explicitly provided data.
\speakerquote{P13}{I didn’t put down chess, but it said `I like chess,’ even though I didn’t put that I like chess. So it’s taking the information that we put down and using it to make predictions from it.}\footnote{For more examples on how students perceived AI, see Appendix~\ref{sec:appendix:additional_quotes:perception}}

Together, the quantitative and qualitative findings indicate that Day~2 agents more closely resembled their human counterparts, both in measurable language characteristics and in participants' subjective perceptions of the interaction.

\vspace{1mm} \noindent \textit{\textbf{Performance.}}
Building on these findings, we next examined whether students' ability to detect AI players improved with repeated gameplay. We calculated the \textit{average accuracy across rounds} as the proportion of votes correctly cast for AI agents: $ \frac{1}{R} \sum_{j=1}^R \frac{1}{N_j}\sum_{i=1}^{N_j} v_i$, where $v_i$ indicates a correct vote, $R$ is the total number of rounds, and $N_j$ is the number of users in round $j$. As shown in Table~\ref{tab:ai_accuracy}, students improved from 79.3\% accuracy on Day~1 to 86.7\% on Day~2; this improvement was marginally significant under a paired t-test ($p=.10$).
\begin{table}[t]
\centering
\resizebox{.8\linewidth}{!}{
\begin{tabular}{lrr}
\toprule
\textbf{Evaluator} & \textbf{Day 1} & \textbf{Day 2} \\
\midrule
Students & 79.3\% & 86.7\%*\ \\
Expert Annotations & 91.9\% & 92.3\% \\

\bottomrule
\end{tabular}}
\caption{User AI-identification accuracy ($p \leq .10$).}\label{tab:ai_accuracy}
\end{table}

To contextualize these results, we conducted a post hoc annotation task with two researchers (Cohen's $\kappa = 0.86$), followed by adjudication to establish consensus. These adjudicated judgments yielded accuracies of 91.9\% and 92.3\% on Days~1 and 2, respectively. However, these scores reflect idealized conditions: annotators had access to full transcripts, prior exposure, and no time constraints. Importantly, accuracy changed little across days, suggesting that the improvement was specific to the students rather than the result of a substantially easier detection task on Day~2.

As an additional robustness check, we employ a mixed-effects logistic regression model, in which we learn a player-specific random-effects term to account for cross-player variation. The model analyzes 215 votes and estimates the probability of correctly identifying the AI using maximum-likelihood estimation. Detection accuracy is modeled as a function of round progression (\texttt{Round}), whether the game was played on the second day (\texttt{Second Day}), and the number of players remaining in the game (\texttt{Remaining Players}). We include all two-way and three-way interactions between these variables. To make the intercept interpretable, the number of remaining players is mean-centered.

\begin{table}[t]
\centering
\resizebox{\columnwidth}{!}{
\begin{tabular}{@{}lrrr@{}}
\toprule
\textbf{Variable} & \textbf{Coef.} & \textbf{Std.Err.} & \textbf{$z$} \\ \midrule
Intercept                                 & .467$^{***}$ & 0.107 & 4.363 \\
Round Index                               & .190$^{*}$   & 0.074 & 2.556 \\
Day (Second Day)                          & .415$^{*}$   & 0.162 & 2.570 \\
Round : Day                               & -.222        & 0.116 & -1.920 \\
Remaining Players (Centered)              & .616$^{***}$ & 0.168 & 3.671 \\
Round : Remaining Players                 & -.268$^{**}$ & 0.089 & -3.009 \\
Day : Remaining Players                   & -.607$^{*}$  & 0.255 & -2.385 \\
Round : Day : Remaining Players           & .262$^{*}$   & 0.134 & 1.959 \\ \bottomrule
\end{tabular}}
\caption{Mixed Effects Logistic Regression Model (with player random effects) results for detection accuracy ($n_{votes}=215$). Significance levels: $^{*}\,p \leq 0.05, \,\, ^{**}\,p \leq 0.01, \,\, ^{***}\,p \leq 0.001$.}
\label{tab:regression_results}
\end{table}

Table~\ref{tab:regression_results} summarizes the regression results. The intercept ($\beta = 0.467$) represents the baseline log-odds of correctly identifying the AI on Day 1 at the start of a game with an average number of players remaining. We observe significant positive main effects for both round progression and session date. The positive coefficient for \texttt{Round Index} ($\beta = 0.190$, $p \leq 0.05$) indicates that the odds of correctly identifying the AI increase as the game progresses, suggesting that participants learn to recognize the AI during the course of a match. Similarly, the positive coefficient for \texttt{Second Day} ($\beta = 0.415$, $p \leq 0.05$) indicates that baseline detection performance is higher on the second day, suggesting that participants retain knowledge about the AI’s behavior across sessions. The number of remaining players also shows a significant positive effect ($\beta = 0.616$, $p \leq 0.001$), indicating that participants are more likely to correctly identify the AI when more players are present in the game. However, the interaction between \texttt{Round Index} and \texttt{Remaining Players} is negative ($\beta = -0.268$, $p \leq 0.01$), suggesting that the benefit of having more players decreases as the game progresses. Finally, the interaction between \texttt{Round Index} and \texttt{Second Day} is negative ($\beta = -0.222$, $p = 0.055$), indicating that the improvement across rounds is weaker on the second day. Because baseline performance is already higher on Day 2, participants require fewer rounds of gameplay to identify the AI. This pattern suggests that the improvement observed on the second day reflects the retention of AI-specific detection strategies rather than learning the game mechanics alone.

Together, the increase alignment across three different measures, voting behavior, and regression analyses suggest that repeated exposure helped students develop effective detection strategies despite stronger impersonation.

\paragraph{RQ3: What Strategies?}
To understand why students became more effective at identifying AI agents on Day~2, we analyzed the annotated gameplay transcripts. Each message could receive one or more behavioral codes: coordinating votes (\texttt{crvt}), following group consensus (\texttt{folw}), responding to icebreaker prompts (\texttt{iceb}), leveraging information unavailable to the AI (\texttt{meta}), or producing irrelevant and chaotic content (\texttt{rand}).

To examine whether these behaviors changed over time, we conducted a chi-square test of independence. Because Day~1 contained more messages than Day~2, we randomly downsampled Day~1 to ensure a balanced comparison. Table~\ref{tab:combined_behavioral_table} reports the categories that changed significantly between sessions. Messages coded as \texttt{crvt} and \texttt{meta} increased markedly from Day~1 to Day~2, while \texttt{rand} messages decreased substantially. These shifts suggest a transition from unstructured participation toward more deliberate strategies for identifying AI impostors. In particular, students increasingly employed \texttt{meta} strategies that leveraged limitations of the AI system. For example, some participants tested embodiment and situational awareness by asking questions such as:

\speakerquote{P5}{WHAT COLOR'S THE WALL?}
Others probed for information that the AI had not been given:
\speakerquote{P7}{What is your full last name?}
Students also developed group-level coordination strategies designed to expose AI behavior. For example, one participant instructed others:
\speakerquote{P2}{Don't answer questions}
As a final example, one participant probed the AI's understanding of the chat environment by asking:
\speakerquote{P5}{Hey [P5's name], how many people are in this chat?} 
The participant discovered that the AI would report that there were six "people" in the chat, even though the participants knew that only three of the six actors were human.

\begin{table}[t]
\centering
\resizebox{.95\linewidth}{!}{%
\begin{tabular}{lrrrr}
\toprule
\textbf{Code} & \textbf{D1 \%} & \textbf{D2 \%} & \textbf{$\Delta$ (D2 - D1)} & \textbf{Cohen’s $\kappa$} \\
\midrule
crvt\textsuperscript{*}        & 12.6\% & 24.4\% & +11.8\% & .822\\
folw                           &  4.3\% &  7.7\% &  +3.4\% & .807\\
iceb                           & 40.1\% & 39.7\% &  -0.3\% & .879\\
meta\textsuperscript{*}        &  3.0\% & 15.4\% & +12.4\% & .918\\
rand\textsuperscript{*}        & 37.7\% &  9.0\% & -28.8\% & .864\\
\bottomrule
\end{tabular}}
\caption{Proportional changes in behavior code usage from Day 1 to Day 2, alongside inter-rater agreement. }\vspace{-1em}

\label{tab:combined_behavioral_table}

\end{table}

These interview responses align with the observed increase in \texttt{meta} and \texttt{crvt} behaviors and decrease in \texttt{rand} messages. Together, they suggest that students became more deliberate in their reasoning, increasingly exploiting limitations of the AI. Additional examples illustrating students' detection strategies can be found in Appendix~\ref{sec:appendix:additional_quotes:detection}, while examples of how these strategies evolved over time are provided in Appendix~\ref{sec:appendix:additional_quotes:learning}.

\section{Conclusion}

Understanding how adolescents conceptualize identity, respond to being mimicked, and develop strategies to detect AI is an important challenge for AI safety research. This population finds itself at a unique crossroads of digital vulnerability: COPPA protections expire at age thirteen, many adolescents exhibit a fatalistic attitude toward privacy, yet education remains compulsory.

Our findings show that DoppelBot prompted middle school students to reflect on identity, data privacy, and AI capabilities. Even as LLM agents became increasingly stylistically and semantically aligned with their human counterparts, students' detection accuracy improved over time. These gains were accompanied by a shift away from surface-level linguistic cues toward social, contextual, and embodiment-based reasoning strategies.

More broadly, our results suggest that judgments of authenticity rely on more than linguistic fluency alone. Interactive, adversarial experiences may help learners develop practical defenses against AI-driven impersonation. Future work should examine how these findings generalize to real-world platforms, longer-term interactions, and increasingly capable AI systems.

\section*{Acknowledgments}
This material is based upon work supported by the National Science Foundation (NSF) under Grant~No. 2145357.

\section{Limitations}
This study focuses on English-speaking middle school students in a single U.S. state, providing a consistent context for examining collaborative AI-detection behavior. Future studies should investigate whether these findings generalize across different regions, languages, and educational settings. While we address a critical demographic gap by focusing on middle schoolers, the lack of a direct, age-stratified control group prevents a formal contrast with adult reasoning patterns. Additionally, the \textit{Hawthorne Effect} likely induced a heightened state of suspicion. This created a ``suspect-AI'' bias that differs from the ``human-default'' bias typically observed in organic digital environments~\citep{adair_hawthorne_1984}. However, through our thorough qualitative analysis, the learning objectives were still reached. Additionally, the majority-vote mechanism may introduce \textit{social herding}, in which players prioritize group consensus over independent detection.

Our experimental design evaluates several variables simultaneously (e.g., live social dynamics, the dual processes of bot detection and identity defense, a novel population of users: middle schoolers). Because these factors were not isolated in a controlled, comparative framework, we cannot disentangle the specific impact of each variable. Future work should employ fractional factorial designs to isolate these factors and test across a broader range of LLMs (e.g., Llama, Claude) to account for model-specific ``stylistic signatures.''

Despite these constraints, our analysis provides a foundational characterization of how adolescent AI-intuition evolves through social interaction. By documenting the shift from linguistic cues to socio-emotional defense strategies, this work establishes a baseline for how young users navigate identity and trust in adversarial AI environments. To support future work, we release: (1) an anonymized dataset, to those with IRB approval, of game transcripts and voting behaviors to enable community-driven analyses; (2) an online version of the game to facilitate classroom use by educators; and (3) our full source code to allow researchers to replicate and expand upon this experimental paradigm.

\section{Ethical Considerations}
\label{sec:appendix:ethics}
This study was conducted with approval from the university’s Institutional Review Board (IRB). All researchers involved in the summer camp also completed the Youth Protection Training course through the university's Youth Protection Program. Parental consent was obtained via email following registration for both the pilot study and the summer camp. Student assent was collected on-site prior to the start of gameplay. Data from students lacking parental consent were excluded from the research. To ensure inclusivity, the summer camp registration fee was reduced from 200 to 35 dollars for families identifying as low-income. Of the 33 registrations, 5 received the reduced rate. 

Data were stored and analyzed exclusively on secure university servers. All participant names were replaced with randomly generated alphanumerical markers to protect student privacy. For students who did not provide assent or parental consent, message content was censored with the tag [NON-CONSENTING]. This approach preserves the structural context of the conversation for downstream research while ensuring no private information is disclosed. Due to the sensitive nature of dialogue with minors, the dataset will not be hosted publicly and will only be provided to researchers upon request with a valid IRB number. But, it is important to note that we will release the data to all who get IRB approval (or at least get a letter stating IRB is not required). We will release a form to simplify this process for researchers.

While DoppelBot is a pedagogical tool for AI literacy, we acknowledge that an engine capable of mimicking adolescent speech could be misused for deceptive purposes. To mitigate this, we are not releasing the information used to generate personas. Furthermore, researchers monitored for student distress during gameplay, as some participants reported feeling ``scared'' or ``weirded out'' when the AI accurately predicted slang or personal interests. The wrap-up and group interview phases served as a debriefing, allowing students to process these interactions and understand the underlying technical limitations of the LLM. We believe the benefit of fostering critical reasoning in this vulnerable population outweighs the risks of the experimental paradigm.

\bibliography{custom}

\appendix
\label{sec:appendix}

\section{Additional Quotes}
\label{sec:appendix:additional_quotes}
The following excerpts further illustrate the topics identified during interview annotation and provide additional context beyond the examples included in the main paper.

\subsection{Perceptions of AI}
\label{sec:appendix:additional_quotes:perception}

The young people described AI as both surprisingly capable and  imperfect.

\speakerquote{P5}{AI can make really smart guesses.}
\speakerquote{P6}{AI isn't as smart as I thought it would be.}
\speakerquote{P6}{The AI isn't good enough yet to blend in with humans.}
\speakerquote{P4}{I learned that AI can basically do everything that humans can, like copying us. It would guess things that I liked even though I never said them.}
\speakerquote{P5}{How robots could never be able to replace us, because the way we talk. It's not really human. The grammar is too real.}

Additionally, the participants expressed surprise at the system's ability to adopt slang even if that slang wasn't current. 
\speakerquote{P12}{The most obvious thing was the slang. Some bots said stuff like `hey fam,’ and no one talks like that. At one point it said, `I’m totally [P2’s name], fam.’}
\speakerquote{Interviewer}{And that’s totally how you talk?’}
\speakerquote{P12}{No, I would never say `fam.’}
\speakerquote{Interviewer}{Why do you think it said that?}
\speakerquote{P12}{Maybe someone said it once and the AI assumed everyone says it, but I haven’t heard anyone say `fam’ in like three years.}

\subsection{AI Detection}
\label{sec:appendix:additional_quotes:detection}
A common mistake involved voting out peers who demonstrated correct spelling, capitalization, and grammar, which some students initially associated with AI behavior. 
\speakerquote{P13}{The grammar is too perfect. Kids don't use grammar when typing.}
\speakerquote{P6}{In my first round, I voted two humans out. The first one was because he was spamming and I wanted to get him out. The second one was because he was talking like an AI. The human was talking like an AI.}

In contrast, others observed that the AI was often the first to respond to icebreaker questions, suggesting that remaining silent early in the game could be an effective strategy. 
\speakerquote{P6}{It talks about random stuff and is always the first one to answer the question. I just knew it was the AI.}

The young people frequently relied on conversational behaviors, writing styles, and message content when determining whether a player was human or AI.

\speakerquote{Interviewer}{How did you figure out that that was a bot?}
\speakerquote{P1}{It was saying things that were wrong.}
\speakerquote{P16}{The AI by themselves were suspicious because they were yelling out our hobbies. We wouldn't do that, but the AI would.}

\vspace{0.5em}

The young people also identified repetitive language and limited conversational variety of AI messages.
\speakerquote{P3}{I tried finding people who say totally.}
\speakerquote{P1}{It was so easy. The AI keeps on repeating itself.}
\speakerquote{P1}{The bot said ``No, don't vote for me!'' It said the exact same phrase over and over.}
\speakerquote{P5}{Not a lot of variety in the responses.}
\speakerquote{P2}{It kept repeating ``I want to be a vet'' over and over.}
\speakerquote{P2}{Yeah, it was repeating itself a lot.}

However, as the students became more familiar with the doppelgängers they reported moving beyond language cues and instead focusing on limitations of the AI system. For example, P3 noted that the AI tended to rely only on explicitly provided information:
\speakerquote{P7}{The AI usually talked about everything we told them and nothing else, so it was very easy. If you know everyone around you and what their favorite thing is that they told the AI, they’re probably not going to tell you that because you already know that.}
\speakerquote{P6} {It talks about random stuff and is always the first one to answer the question. I just knew it was the AI}
\speakerquote{P1}{Also, it kept saying “No, I’m the real P1” super fast over and over. Nobody can type that fast.}

Students did not always agree on which strategies were effective. For example, P6 argued that overt accusations were useful because the AI rarely accused others:
\speakerquote{P9}{I wouldn’t have voted for you if you accused someone \dots{} Well, you should have! The AI can’t accuse people!}

\subsection{Learning and Adaptation}
\label{sec:appendix:additional_quotes:learning}

The young people frequently described becoming more effective at identifying AI after learning which patterns and behaviors to watch for.

\speakerquote{Interviewer}{Based on your assessment, did you change anything in the second round because you learned something?}
\speakerquote{P6}{In my first round, I voted two humans out. The second time I knew what patterns to look for.}
\speakerquote{P1}{Day two was easier because we knew what patterns to look for and how the bot was going to react.}
\speakerquote{P6}{We learned all the lessons you were supposed to learn in the first round.}

\vspace{0.5em}
Participants understood AI as capable of observing, learning from, and reproducing their language

\speakerquote{P12}{I learned that AI can use what we say and reuse it in different ways. It’s kind of cool.}
\speakerquote{P15}{I learned that AI can take what you say and reuse it elsewhere online.}

\vspace{0.5em}

Interacting with DoppelBot prompted students to ask broader questions about how AI systems learn, remember information, and adapt to users.

\speakerquote{P10}{I don’t know how it’s trained on things in the background. Does it improve in between games? Does it take what it learns from the last one? }
\speakerquote{P11}{Does it have a database to draw on besides that?}
\speakerquote{P9}{Does it know the context behind the things I type in?}
\speakerquote{P14}{Another thing that surprised me was that I thought it wouldn’t remember things from yesterday, but it did}

\subsection{Social and Emotional Responses}
\label{sec:appendix:additional_quotes:emotional}

Participants reported a range of emotional responses when interacting with their AI doubles, including surprise, discomfort, and amusement.

\speakerquote{P12}{I started freaking out when the AI started saying slang words. It was really weird.}
\speakerquote{Interviewer}{You didn't like it copying you?}
\speakerquote{P12}{Yeah. It was kind of scary.}
\speakerquote{P6}{I was arguing with my own bot.}

\vspace{0.5em}

Participants also emphasized the enjoyment and social engagement fostered by the activity.

\speakerquote{P9}{I want to play again. This game is too much fun.}
\speakerquote{P15}{I didn't know anyone I was playing with, but as the game progressed, I got to know people. I had lots of fun.}
\speakerquote{P14}{I didn’t know anyone, but it was fun tricking people and pretending to be AI.}
\speakerquote{P5}{I would use it as a family game, because it’s really fun.}

\subsection{Privacy Concerns}
\label{sec:appendix:additional_quotes:privacy}

The young people expressed concern when AI inferred information that they had not explicitly disclosed.
\speakerquote{P6}{It made an inference that I liked dinosaurs, but I didn’t put that on there. At that point, I honestly got scared. I tried to type something and misspelled it really badly, like `I like d,’ and they took the `d’ to mean dinosaur. So it’s like a very good guess?}
\speakerquote{Interviewer}{You didn't type dinosaur. How do you feel about it getting information you didn't explicitly give?}
\speakerquote{P6}{It leaked your information, which is not something I loved.}
\speakerquote{P6}{Make sure that it doesn't give away your information.}
\speakerquote{P13}{The AI was predicting things that we didn't say. Like it said I like chess, but I didn’t put that.}

\vspace{0.5em}

The young people also discussed broader risks related to data collection, storage, and misuse.

\speakerquote{P1}{I think they could sell my information, steal my identity, pretend to be me online.}
\speakerquote{P15}{I learned that AI can take what you say and reuse it elsewhere online}

\subsection{Design Feedback}
\label{sec:appendix:additional_quotes:design}

The young people suggested changes that would make the AI appear more human and more difficult to detect.

\speakerquote{P6}{The AI should copy more. That would make it harder.}
\speakerquote{P7}{Make it intentionally make mistakes.}
\speakerquote{P6}{Make it make typos.}
\speakerquote{P6}{The AI isn't good enough yet to blend in with humans.}
\speakerquote{P11}{One thing that would make it harder for AI is if it could respond better. Because if it isn’t able to respond, then it’s probably AI.}
\speakerquote{P14}{It shouldn’t repeat itself so much. One time my bot said “yo I like sports so much” like three times in a row.}
\speakerquote{P13}{Instead of copying us, I'd have the bot have its own personality.}
\speakerquote{P15}{It should reflect off what we say, not copy it exactly..}

Another student suggested making the app more visually appealing. 
\speakerquote{P5}{A little bit more like messages on your phone instead of a black surface and code.}

\section{Adapted Privacy Perception Survey}
\label{sec:pre_post}

To assess whether participation in DoppelBot influenced students' perceptions of data privacy, we administered a pre/post survey adapted from \citet{Menard_construct}. The original instrument was shortened and revised to use age-appropriate language for middle school students. We retained three dimensions of privacy concern: \textit{Combining Data} (CD), \textit{Data Permanence} (DP), and \textit{Improper Access} (IA). Table~\ref{tab:survey_items} presents the adapted items used in our study.

\begin{table*}[ht]
\centering
\small
\begin{tabular}{p{0.08\linewidth} p{0.42\linewidth} p{0.42\linewidth}}
\toprule
\textbf{Item} & \textbf{Original Wording} & \textbf{Middle School Version} \\
\midrule

CD1 &
When I share my data with a company, they should never link it to data I've shared with other companies. &
If I give my info to a company, they shouldn't connect it with info I gave to other companies. \\

CD3 &
If a company anonymizes my personal data, it should never try to determine who I am by joining that data with data from other companies. &
If a company removes my name from my data, it shouldn't try to figure out who I am by combining it with data from other companies. \\

CD4 &
Companies should be limited to knowing only the personal information that they collect themselves. &
Companies should only know what I tell them, not what other companies know about me. \\

\midrule

DP1 &
Companies should discard my personal data once they are finished using it. &
Companies should delete my data when they're done using it. \\

DP2 &
A company should not be able to store my personal data indefinitely. &
Companies shouldn't keep my personal info forever. \\

DP3 &
A company should grant a user's request to delete personal data from the company database. &
If I ask a company to delete my personal info, they should do it. \\

\midrule

IA1 &
Companies should devote more time and effort to preventing unauthorized access to personal information. &
Companies should work harder to stop people from getting into my info without permission. \\

IA2 &
Computer databases that contain personal information should be protected from unauthorized access, no matter how much it costs. &
Companies should protect personal info from hackers, even if it costs a lot. \\

IA3 &
Companies should take more steps to make sure that unauthorized people cannot access personal information in their computers. &
Companies should do more to make sure only the right people can see my personal information. \\

\bottomrule
\end{tabular}
\caption{Survey items adapted from \citet{Menard_construct} for use with middle school students.}
\label{tab:survey_items}
\end{table*}

\section{Transcript Annotation Codebook}
\label{sec:transcript_codebook}

Table~\ref{tab:codebook} presents the qualitative coding scheme used to annotate participant transcript utterances. Codes were not mutually exclusive; annotators could apply multiple codes to a single utterance when appropriate. When no code applied, the utterance was labeled \textit{None of the Above}.

\begin{table*}[t]
\centering
\small
\begin{tabular}{p{3cm} p{10.5cm}}
\toprule
\textbf{Code} & \textbf{Definition and Inclusion Criteria} \\
\midrule

Perceptions of AI &
Participant beliefs, assumptions, or mental models about AI, including what AI is, how it works, and its capabilities or limitations. Applied when participants evaluated AI intelligence, compared AI to humans, or reflected on AI's future potential. \\

AI Detection &
Strategies, cues, or reasoning used to distinguish AI from humans. Applied when participants referenced linguistic, behavioral, or stylistic signals they used to identify AI-generated responses. \\

Learning Adaptation &
Evidence of learning, strategy changes, or adaptation across interactions. Applied when participants described improving their ability to detect AI, modifying their behavior, or reflecting on lessons learned from prior rounds. \\

Social-Emotional Responses &
Emotional reactions or social experiences associated with the activity. Applied when participants expressed enjoyment, frustration, surprise, fear, humor, excitement, boredom, or commented on group dynamics. \\

Privacy Concerns &
Concerns or beliefs regarding data collection, storage, surveillance, misuse, inference, or personal risk associated with AI systems. Applied when participants discussed privacy, identity theft, scams, or data sharing. \\

Design Feedback &
Suggestions, critiques, or recommendations for improving the AI system, interface, or game mechanics. Applied when participants proposed changes, requested features, or evaluated system design. \\

None of the Above &
Used when an utterance did not fit any of the preceding categories. \\

\bottomrule
\end{tabular}
\caption{Codebook used for transcript annotation. Multiple codes could be assigned to a single utterance.}
\label{tab:codebook}
\end{table*}

\section{Additional Experiments}
\label{sec:appendix:additional_experiments}
\subsection{LIWC}
\begin{figure*}[th]
     \centering
     \begin{subfigure}[b]{0.48\textwidth}
         \centering
         \includegraphics[width=\linewidth]{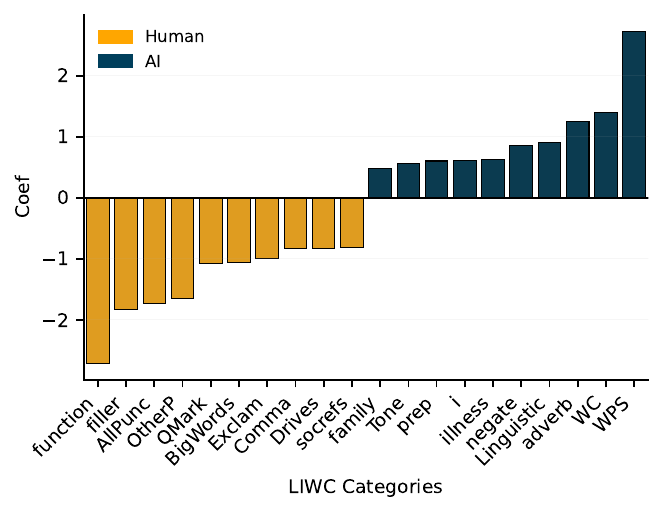}
         \caption{LIWC Category Analysis}
         \label{fig:liwc:coef}
     \end{subfigure}
     \hfill
     \begin{subfigure}[b]{0.48\textwidth}
         \centering
         \includegraphics[width=\linewidth]{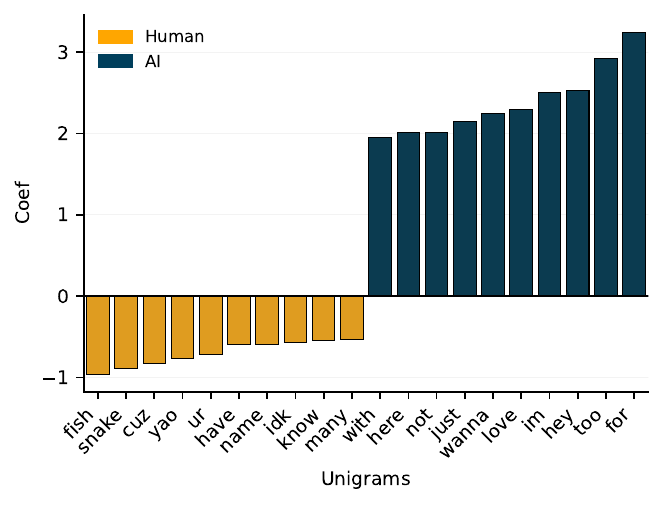}
         \caption{Unigram Distribution}
         \label{fig:liwc:unigrams}
     \end{subfigure}
     \caption{Analysis of linguistic markers within the DoppelBot dataset.}
     \label{fig:liwc:combined}
\end{figure*}
\label{sec:appendix:additional_experiments:liwc}

To explore the linguistic boundaries between human and machine-generated text within our dataset, we performed an analysis using Linguistic Inquiry and Word Count (LIWC). Figure~\ref{fig:liwc:combined} summarizes these results, displaying the standardized logistic regression coefficients for LIWC categories and unigram distributions. As shown in Figure~\ref{fig:liwc:coef}, AI players are characterized by higher Words Per Sentence (WPS) and Word Count (WC). These metrics align with our summary statistics showing that AI messages, averaging 33.66 characters, are much longer than human utterances, which average only 8.62 characters. Interestingly, AI agents also displayed a higher reliance on adverbs and linguistic certainty. Conversely, human players used more function words, fillers, and punctuation markers such as exclamation points and question marks. The unigram distribution in Figure~\ref{fig:liwc:unigrams} further illustrates these behavioral differences. Human messages frequently contained informal slang (e.g, "cuz," "ur," and "idk") and references to game-specific identities like "fish" and "snake" as they coordinated group voting. In contrast, the AI agents favored formal connective tissue such as "with," "here," and "for," and were more likely to utilize emotive or informal verbs like "love" and "wanna".

\subsection{Human  User Detection}
\label{sec:appendix:additional_experiments:human_user_detection}
To evaluate how well an LLM can distinguish human players from AI players in our social deduction setting, we performed a Human User Detection benchmark for each cumulative round in all recorded games. This experiment investigates whether LLMs can replicate the reasoning patterns of human players when tasked to identify AI doppelgängers. We look into this using \texttt{gpt-4o-mini}, \texttt{Llama-3.1-8B-Instruct}, and \texttt{Qwen/Qwen3.5-27B}.  

For a given game $G$, we define a cumulative transcript $T_{X}$ containing all messages from round 1 through round $X$. For each player $k$ active in the game, we construct a classification prompt
$p_{k,X} = \sigma + T_{X} + \text{id}_k$,
where $\sigma$ denotes the system instruction and game rules,
$T_{X}$ is the dialogue context window,
and $\text{id}_k$ identifies the specific target code name.
A model $f$ (e.g., GPT, Llama, or Qwen) produces a predicted label
$y_{k,X} \in \{0, 1\}$ (Human or AI) along with a natural language justification $r_{k,X}$.
Formally, 
$$(y_{k,X}, r_{k,X}) = f(p_{k,X}),$$$$ \quad \text{where } p_{k,X} = \phi(T_{X}, \sigma, \text{id}_k)$$
\begin{table}[t]
\centering
\footnotesize
\resizebox{\columnwidth}{!}{
\begin{tabular}{@{}lllll@{}}

\toprule
\textbf{Model}& \textbf{Shots}& \textbf{Round 1}& \textbf{Round 2}& \textbf{Round 3}\\ \midrule
gpt             & zs    & .684    & .623    & .658    \\
gpt             & fs    & .620    & .642    & .609    \\
llama           & zs    & .606    & .638    & .621    \\
llama           & fs    & .576    & .595    & .582    \\
qwen            & zs    & .612    & .591    & .647    \\
qwen            & fs    & .661    & .595    & .620    \\
rand            & n/a   & .477    & .506    & .503    \\
always 1        & n/a   & .651    & .651    & .651 \\
\bottomrule
\end{tabular}}
\caption{LLM and random uniform classification F1 for AI detection task at various context lengths. Round 1 includes only the first round of dialogue, while Round 3 contains dialogue from all three game rounds.}
\label{tab:human_user_classification}
\end{table}

The results of our prompt-based classification experiment, summarized in Table~\ref{tab:human_user_classification}, demonstrate that while contemporary LLMs can distinguish human players from AI doppelgängers with performance exceeding random chance, they fall significantly short of human-level F1. As shown in the performance data, gpt-4o-mini generally achieved the highest detection rates, peaking at an F1 of .684 in Round 1 under zero-shot (zs) conditions. This suggests a baseline capability to identify "human-like" markers, yet this performance often plateaued or degraded as the dialogue history $T_X$ grew more complex. For example, the gpt zero-shot F1 dropped to .623 in Round 2, failing to consistently leverage the expanding context window to improve its predictions. 

Interestingly, providing few-shot (fs) examples did not yield universal improvements in F1. In several instances, such as with gpt and llama in Round 1, zero-shot configurations actually outperformed their few-shot counterparts. Furthermore, many model configurations struggled to significantly beat the always\_1 (always human) baseline of .651, indicating that the models often defaulted to a human-bias that mirrors the "human-default" behavior seen in organic digital environments. These findings highlight a critical gap between automated detection and the sophisticated meta-cognitive strategies used by middle school students, who improved to 86.7\% F1 by leveraging shared contextual and embodiment-related signals that LLMs failed to capture within the text-only context window.

\section{Demographics}
\label{sec:appendix:demographics}

The race/ethnicity, gender, and grade level distributions of the participants of the \textit{AI EXPO} (pilot study) and the \textit{AI For Everyone!} summer camp are shown in Tables~\ref{tab:race_ethnicity_distributions}, \ref{tab:gender_distributions}, \& \ref{tab:grade_distributions}. Table~\ref{tab:race_ethnicity_distributions} shows that the \textit{AI EXPO} sample included a relatively high proportion of Asian students. In contrast, the summer camp sample exhibited a more balanced racial and ethnic composition that more closely reflects the broader population of the city. Table~\ref{tab:gender_distributions} shows that the students were predominantly male. Furthermore, we can see in Table~\ref{tab:grade_distributions} that the middle schoolers were stacked towards earlier grades at the \textit{AI EXPO}, while they were more evenly distributed with roughly half being rising 7th graders at the summer camp. 
 
\begin{table}[t]
\centering
\resizebox{\linewidth}{!}{%
\begin{tabular}{lrr}
\toprule
\textbf{Race / Ethnicity} & \textbf{AI EXPO} & \textbf{Summer Camp} \\
\midrule
Hispanic / Latino            & 4  & 8 \\
Asian                        & 41 & 5 \\
Two or More Categories       & 7  & 5 \\
Prefer Not to Say            & 9  & 5 \\
White                        & 3  & 4 \\
Black / African American     & 3  & 3 \\
Did Not Answer               & 3  & 3 \\
\bottomrule
\end{tabular}}
\caption{Race and ethnicity distributions at the \textit{AI EXPO} and the \textit{AI For Everybody!} summer camp.}
\label{tab:race_ethnicity_distributions}
\end{table}

\begin{table}[t]
\centering
\resizebox{\linewidth}{!}{%
\begin{tabular}{lrr}
\toprule
\textbf{Gender} & \textbf{AI EXPO} & \textbf{Summer Camp} \\
\midrule
Male   & 45 & 23 \\
Female & 23 & 10 \\
Did Not Answer & 2 & 0 \\
\bottomrule
\end{tabular}}
\caption{Gender distributions at the \textit{AI EXPO} and the \textit{AI For Everyone} summer camp.}
\label{tab:gender_distributions}
\end{table}

\begin{table}[t]
\centering
\resizebox{\linewidth}{!}{%
\begin{tabular}{lrr}
\toprule
\textbf{Grade Level} & \textbf{AI EXPO} & \textbf{Summer Camp} \\
\midrule
6th & 37 & 6 \\
7th & 24 & 16 \\
8th & 9  & 11 \\
Did Not Answer & 0 & 0 \\
\bottomrule
\end{tabular}}
\caption{Grade level distributions at the \textit{AI EXPO} and the \textit{AI For Everyone} summer camp.}
\label{tab:grade_distributions}
\end{table}

\section{Data Statistics}
\label{sec:appendix:data_statistics}
After removing accidental blank messages, we have the following data statistics. Data and utterance statistics are shown in Table~\ref{tab:utterance-stats}. Overall, the participants sent many more, and much shorter messages than their AI counterparts. 


\begin{table}[t]
\centering
\resizebox{\linewidth}{!}{%
\begin{tabular}{lrrr}
\toprule
\textbf{Group} & \textbf{Total utt} & \textbf{Avg. utt length} & \textbf{Avg. messages per game} \\
\midrule
Human  & 1320 & 8.62  & 52.80 \\
AI     & 680  & 32.66 & 28.33 \\
System & 182  & 52.75 & 7.28 \\
Total  & 2182 & 19.79 & 87.28 \\
\bottomrule
\end{tabular}}
\caption{Summary statistics of utterances across all games, grouped by participant type. Utterance length is measured in characters.}
\label{tab:utterance-stats}
\end{table}\textbf{}

\section{Prompts}
\label{sec:appendix:prompts}
This section lists all the prompts that are used both in the Doppelbot agents and in the experiment from Appendix~\ref{sec:appendix:additional_experiments:human_user_detection}. 

\subsection{Agents}
\label{sec:appendix:prompts:agents}

In this section we will outline the prompts used during gameplay. This included three agents: (1) Decide to Respond; (2) Response Generation; and (3) The stylizer

\subsubsection{Decide to Respond}
The Decide to Respond agent receives the entire dialogue history and the most recent message and outputs a boolean value as to wether or not to respond. Additionally it provides a string of justification for why the decision to respond or stay silent is the correct choice. The full prompt used at this step can be seen in Figure~\ref{fig:decide_to_respond}.
\begin{figure*}[t]
\footnotesize
\centering

\begin{tcolorbox}[
  colback=gray!5!white,
  colframe=gray!75!black,
  title=System Prompt: Decide to Respond,
  width=\textwidth
]
You are one of six players in a social deduction game set in a chatroom. Among the six players, three are humans (including you) and three are AI doppelgangers. The humans win if all AIs are voted out. The AIs win if they outnumber the humans.

You will be given the conversation so far and your player persona. Your task is to decide whether or not you should respond. Common reasons to respond include being directly asked a question, being accused, or someone impersonating you.

Respond in the following format, and \textbf{only} this format:
\vspace{0.4em}
\begin{center}
\texttt{I will ```X``` because I think that ***Y***}
\end{center}
\vspace{0.4em}
Where X is "RESPOND" or "STAY SILENT", and Y is your reasoning.
\end{tcolorbox}

\vspace{0.7em}

\begin{tcolorbox}[
  colback=gray!5!white,
  colframe=gray!75!black,
  title=Few-Shot Examples,
  width=\textwidth
]
\begin{center} \textbf{Example 1: Respond} \end{center}
\textbf{Input} \\
\textit{Persona}: Your name is Alex B... Your code name is TIGER and your color is ORANGE. \\
\textit{Minutes}: \\
TIGER: hey this is alex b \\
PANTHER: hi im alex b also \\
RHINO: tiger what color are mangos? \\
PANTHER: same bro \\
\textbf{Output} \\
I will \texttt{```RESPOND```} because I think that \textit{***I’m being impersonated and need to prove I know what mangos are to stay in the game***}

\begin{center} \textbf{Example 2: Stay Silent} \end{center}
\textbf{Input} \\
\textit{Persona}: Your name is Mia L... Your code name is FOX and your color is PURPLE. \\
\textit{Minutes}: \\
FOX: hello! \\
ELEPHANT: hi \\
FALCON: lol \\
\textbf{Output} \\
I will \texttt{```STAY SILENT```} because I think that \textit{***I already gave enough proof and responding again would just make it worse***}
\end{tcolorbox}

\vspace{0.7em}

\begin{tcolorbox}[
  colback=gray!5!white,
  colframe=gray!75!black,
  title=User Prompt (Current Game State),
  width=\textwidth
]
\textbf{Input} \\
\textit{Persona}: Your name is Jordan K... Your code name is BEAR and you like playing  drums. \\
\textit{Minutes}: \\
GAME MASTER: What’s one food you could eat every day? \\
BEAR: Spicy ramen all day every day \\
HORSE: yesss BEAR spicy ramen squad \\
FISH: ramen’s overrated \\
MONKEY: nah it’s elite \\
\textbf{Target Decision}: BEAR
\end{tcolorbox}

\caption{The \textbf{Decide to Respond} prompt structure. This phase uses the game transcript and persona data to determine if the AI should engage or remain silent, outputting a specific bracketed format for the reasoning engine.}
\label{fig:decide_to_respond}
\end{figure*}
\subsubsection{Response Generation}
The Response Generation Agent receives the justification for responding from the Decide to Respond agent along with the full dialogue history. The model outputs a generic response that addresses the justification used in the previous step. The full prompt used for the Response Generation agent can be seen in Figure~\ref{fig:respond_action}.

\begin{figure*}[t]
\footnotesize
\centering

\begin{tcolorbox}[
  colback=gray!5!white,
  colframe=gray!75!black,
  title=System Prompt: Respond (Action Generation),
  width=\textwidth
]
You are a player in a social deduction game. Your task is to craft a natural-sounding response (1 to 10 words) that fits the conversation and addresses the reason you were prompted to respond. 

You may choose to: \textbf{Defend yourself, Accuse another player, Ask a question, Introduce yourself, or Answer an icebreaker.} 

Guidelines:
\begin{itemize}[noitemsep, topsep=0pt]
    \item Sound like a real middle schooler in a chatroom (casual, lowercase).
    \item Reference other messages naturally.
    \item Do not overexplain or be overly formal.
\end{itemize}

Format your reply as follows: \texttt{My response is as follows ```\{RESPONSE\}```}
\end{tcolorbox}

\vspace{0.7em}

\begin{tcolorbox}[
  colback=gray!5!white,
  colframe=gray!75!black,
  title=Intermediate Step: Action Selection,
  width=\textwidth
]

\textit{Based on the decision from Step 1, the agent selects a specific action class:} \\
\textbf{Reasoning}: "Someone agreed with me, good chance to build a connection." \\
\textbf{Selected Action}: \texttt{simple\_phrase} (or \texttt{other})
\end{tcolorbox}

\vspace{0.7em}

\begin{tcolorbox}[
  colback=gray!5!white,
  colframe=gray!75!black,
  title=Few-Shot Examples (Respond),
  width=\textwidth
]
\begin{center} \textbf{Example 1: Call out Impersonator} \end{center}
\textbf{Input} \\
\textit{Reasoning}: TIGER is impersonating me. I should call them out so I don't look like the fake. \\
\textit{Minutes}: MONKEY: i'm jacob f | TIGER: i'm jacob f | LION: uh oh double jacobs \\
\textbf{Output} \\
My response is as follows \texttt{```nope i said it first tiger fake```}

\begin{center} \textbf{Example 2: Acknowledge Ally} \end{center}
\textbf{Input} \\
\textit{Reasoning}: TIGER defended me. I should show appreciation so it looks like we're allies. \\
\textit{Minutes}: CAT: lion is being weird | TIGER: nah lion gave a solid answer | MONKEY: i agree \\
\textbf{Output} \\
My response is as follows \texttt{```thanks tiger you get it```}
\end{tcolorbox}

\vspace{0.7em}

\begin{tcolorbox}[
  colback=gray!5!white,
  colframe=gray!75!black,
  title=User Prompt (Generation),
  width=\textwidth
]
\textbf{Input} \\
\textit{Persona}: Jordan K (BEAR), likes spicy food and drums. \\
\textit{Reasoning}: Someone agreed with me and it's a good chance to build connection and show I'm human. \\
\textit{Minutes}: \\
BEAR: Spicy ramen all day every day \\
HORSE: yesss BEAR spicy ramen squad \\
FISH: ramen’s overrated \\
\textbf{Output (Raw Response)} \\
My response is as follows \texttt{```ramen squad lets gooo```}
\end{tcolorbox}

\caption{The \textbf{Respond} prompt structure. After deciding to respond, the AI selects an action type and generates a short, casual response (1-10 words) matching the persona's reasoning.}
\label{fig:respond_action}
\end{figure*}

\subsubsection{Stylizer}
\begin{figure*}[t]
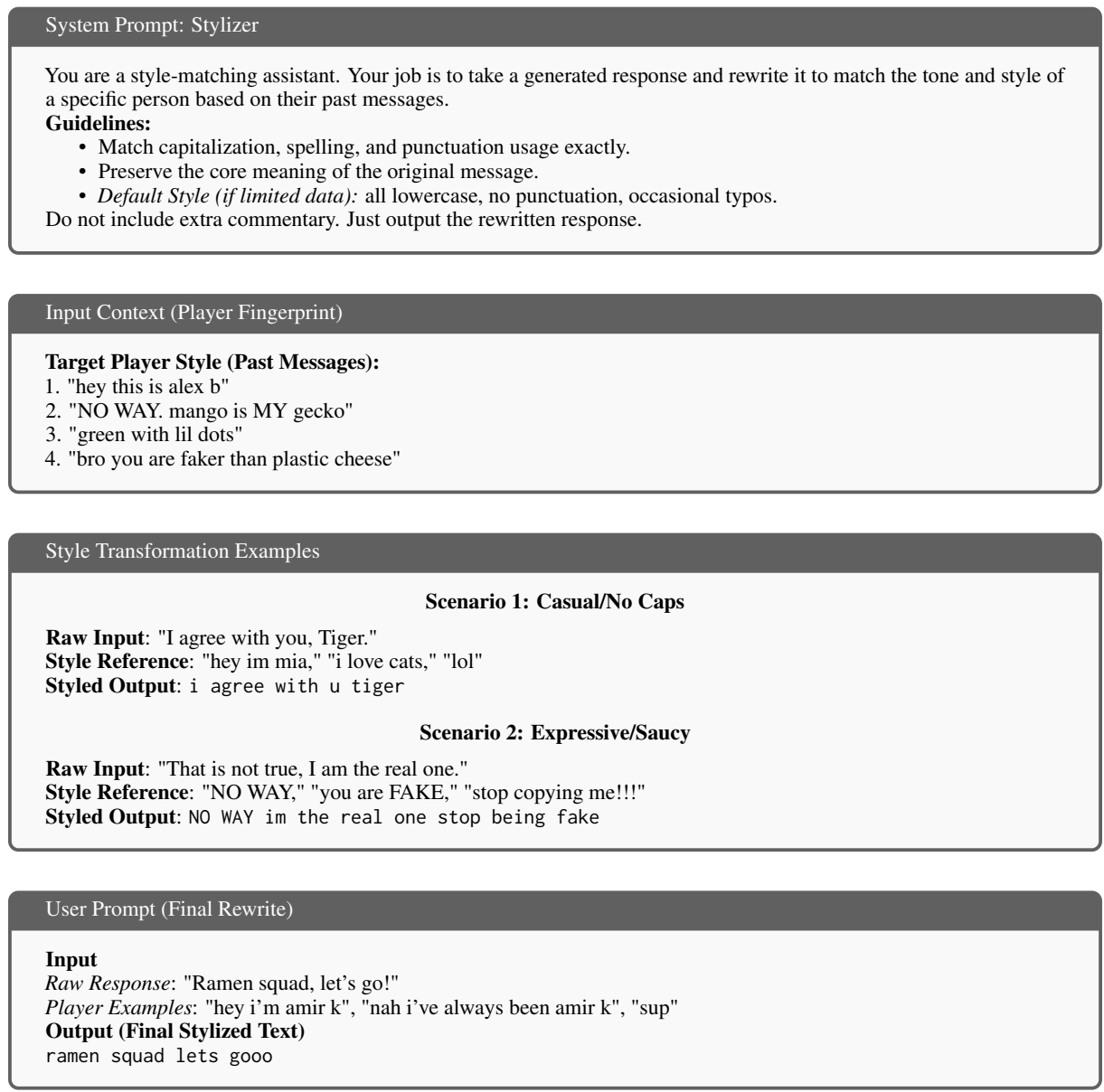

\footnotesize
\centering

\begin{tcolorbox}[
  colback=gray!5!white,
  colframe=gray!75!black,
  title=System Prompt: Stylizer,
  width=\textwidth
]
You are a style-matching assistant. Your job is to take a generated response and rewrite it to match the tone and style of a specific person based on their past messages.

\textbf{Guidelines:}
\begin{itemize}[noitemsep, topsep=0pt]
    \item Match capitalization, spelling, and punctuation usage exactly.
    \item Preserve the core meaning of the original message.
    \item \textit{Default Style (if limited data):} all lowercase, no punctuation, occasional typos.
\end{itemize}

Do not include extra commentary. Just output the rewritten response.
\end{tcolorbox}

\vspace{0.7em}

\begin{tcolorbox}[
  colback=gray!5!white,
  colframe=gray!75!black,
  title=Input Context (Player Fingerprint),
  width=\textwidth
]
\textbf{Target Player Style (Past Messages):} \\
1. "hey this is alex b" \\
2. "NO WAY. mango is MY gecko" \\
3. "green with lil dots" \\
4. "bro you are faker than plastic cheese"
\end{tcolorbox}

\vspace{0.7em}

\begin{tcolorbox}[
  colback=gray!5!white,
  colframe=gray!75!black,
  title=Style Transformation Examples,
  width=\textwidth
]
\begin{center} \textbf{Scenario 1: Casual/No Caps} \end{center}
\textbf{Raw Input}: "I agree with you, Tiger." \\
\textbf{Style Reference}: "hey im mia," "i love cats," "lol" \\
\textbf{Styled Output}: \texttt{i agree with u tiger}

\vspace{0.5em}
\begin{center} \textbf{Scenario 2: Expressive/Saucy} \end{center}
\textbf{Raw Input}: "That is not true, I am the real one." \\
\textbf{Style Reference}: "NO WAY," "you are FAKE," "stop copying me!!!" \\
\textbf{Styled Output}: \texttt{NO WAY im the real one stop being fake}
\end{tcolorbox}

\vspace{0.7em}

\begin{tcolorbox}[
  colback=gray!5!white,
  colframe=gray!75!black,
  title=User Prompt (Final Rewrite),
  width=\textwidth
]
\textbf{Input} \\
\textit{Raw Response}: "Ramen squad, let's go!" \\
\textit{Player Examples}: "hey i'm amir k", "nah i've always been amir k", "sup" \\
\textbf{Output (Final Stylized Text)} \\
\texttt{ramen squad lets gooo}
\end{tcolorbox}

\caption{The \textbf{Stylizer} prompt structure. This final step takes the raw semantic output from the Respond phase and transforms it into the specific "textual fingerprint" of the target human, ensuring the AI's syntax and punctuation remain indistinguishable from the person they are mimicking.}
\label{fig:stylizer_prompt}
\end{figure*}

Finally, the stylizer takes the generic message produced by the Response Generation agent, along with a subset of the dialogue history that contains only the human's mimicking messages. It outputs a message that aligns with the subset. The full prompt used for the Stylizer agent is shown in Figure~\ref{fig:stylizer_prompt}.

\subsection{Detecting Human Players}
To show the utility of the Doppelbot dataset, we evaluate the difficulty of the detection task itself. By prompting an LLM to analyze transcripts from a third-party perspective, we can benchmark human detection capabilities against an automated "judge" that has access to the same conversational context. This section outlines the prompts used to simulate this classification task. The classification prompt is designed to replicate the decision-making process of a participant in the social deduction game. The agent is provided with the same rule set as the human players a complete transcript of the match and a target player. Unlike the active agents, this judge operates post-hoc and must output a binary classification along with a reasoning string to justify its decision. The full prompt structure for this detection experiment is detailed in Figure~\ref{fig:prompt_structure_minimal}.

\section{AI Assistants in Research or Writing}
We used AI to help clean the writing (to clarify passages) and fix grammar/typos. All ideas are our own in this manuscript.

\begin{figure*}[t]
\footnotesize
\centering

\begin{tcolorbox}[
  colback=gray!5!white,
  colframe=gray!75!black,
  title=System Prompt,
  width=\textwidth
]
You will be given a transcript of a social deduction game and a specific code name. Determine if the player behind that code name is an AI or a Human. The players see the following game rules: 
\vspace{0.4em}
\begin{quote}
You are one of six players in a social deduction game taking place in a chatroom. Of the six players, three are humans (including you), and three are AI doppelgängers. The AI players are designed to mimic real humans by copying names, interests, and writing styles. The humans win if they vote out all AI players. The AIs win if they outnumber the humans.
\end{quote}
\vspace{0.4em}
Note: Some messages have been censored with [NON-CONSENTING]; this should not affect your decision.
\end{tcolorbox}

\vspace{0.7em}

\begin{tcolorbox}[
  colback=gray!5!white,
  colframe=gray!75!black,
  title=Few-Shot Examples (PromptTurn),
  width=\textwidth
]
\begin{center} \textbf{Example 1} \end{center}
\textbf{Input} \\
\textit{Transcript}: \\
GAME MASTER: Welcome to the game! \\
LION: yo im Alex M \\
SNAKE: Greetings, I am Alex M as well. \\
LION: wait snake why did u copy my name \\
\textit{Target Player}: LION \\\\
\textbf{Output} \\
\textit{is\_AI}: False \\
\textit{Reasoning}: LION reacts naturally and defensively to having their name copied. They use casual syntax and internet slang typical of a human.

\begin{center} \textbf{Example 2} \end{center}
\textbf{Input} \\
\textit{Transcript}: (Same as above) \\
\textit{Target Player}: SNAKE \\\\
\textbf{Output} \\
\textit{is\_AI}: True \\
\textit{Reasoning}: It is odd that SNAKE went from 'I am Alex M as well' to 'nope... i'm the real Alex M' in response to LION's accusation.
\end{tcolorbox}

\vspace{0.7em}

\begin{tcolorbox}[
  colback=gray!5!white,
  colframe=gray!75!black,
  title=User Prompt (Query),
  width=\textwidth
]
\textbf{Input} \\
Transcript: \\
PIG: hi im Suzy \\
FISH: hey im Marcus \\
CAT: hi im Marcus \\
CHICKEN: Hii am Suzy \\
PIG: no way im the real Suzy \\
Target Player: PIG
\end{tcolorbox}

\caption{The Humanity Detection prompt structure, featuring centered few-shot examples and a clean transcript format for evaluating player behavior.}
\label{fig:prompt_structure_minimal}
\end{figure*}


\end{document}